\documentclass[letterpaper]{article}
\usepackage[preprint]{aaai2027}
\usepackage[hyphens]{url}
\usepackage{graphicx}
\usepackage{natbib}
\usepackage{amsmath}
\usepackage{amssymb}
\usepackage{array}
\usepackage{booktabs}
\usepackage{multirow}
\usepackage{pdfpages}

\title{STAIR: Semantic-Temporal Automaton for Interpretable Reasoning \\ in Temporal Question Answering}
\author{Xinlong Dai, Jinchuan Zhang, Lei Gao, Xinzhe Hu, Yuefeng He, Hui Gao}
\affiliations{
University of Electronic Science and Technology of China (UESTC)\\
Correspondence: \texttt{jc.zhang@uestc.edu.cn}
}

\begin{document}
\maketitle

\begin{abstract}
By leveraging large-scale pretraining, LLMs can interpret diverse temporal expressions and question formulations without task-specific training. However, existing prompt-based neuro-symbolic systems continue to rely on LLMs for both semantic interpretation and exact temporal inference. Consequently, discrete decisions regarding intervals, time anchors, and ordered states remain vulnerable to probabilistic errors and difficult to verify. We present STAIR, a \textbf{S}emantic-\textbf{T}emporal \textbf{A}utomaton for \textbf{I}nterpretable \textbf{R}easoning. STAIR separates semantic interpretation from precise temporal inference: an answer-free LLM adapter maps complex question formulations to normalized temporal intents, while a deterministic temporal automaton with finite control and guarded transitions executes the corresponding policies over canonicalized evidence. Following a rule-first design, STAIR resolves standard questions without invoking an LLM and applies semantic adaptation only when the rule path fails to produce an executable intent. This approach reduces free-form reasoning, making temporal decisions verifiable and interpretable. Specifically, guarded execution supports precise point-time containment and before/after selection, while semantic adaptation handles non-exact intervals and time-anchored queries. Across the TimeQA-Easy, TimeQA-Hard, TempReason-L2, and TempReason-L3 datasets, STAIR consistently outperforms strong baselines in the TQA task using matched model settings, achieving average F1 improvements of 16.57\% and 3.10\% when utilizing the Qwen2.5-7B and GPT-4o-mini models, respectively. Furthermore, ablations and diagnostic analyses demonstrate that STAIR excels at handling both boundary-sensitive and order-sensitive queries, while its guarded execution and semantic adaptation ensure precise point-time reasoning and inexact intervals, respectively.
\end{abstract}

\section{Introduction}
Temporal question answering (TQA) requires models to answer questions over time-indexed evidence. To provide reliable and high-confidence answers, the reasoning process requires both the semantic interpretation of diverse question formulations and precise, discrete selection over temporally ordered evidence.

TQA datasets generally share an input-output structure consisting of a question $Q$, a temporal context $C$, and an answer $A$. The temporal context frequently contains multiple time-indexed facts that can be normalized as $(s,r,o,t^s,t^e)$. Even within the same context $C$, different questions $Q$ may impose distinct temporal constraints, requiring boundary-aligned interval matching, non-exact interval overlap, point-in-interval containment, or before/after selection relative to temporal or entity anchors. These operations expose two recurring challenges: boundary-sensitive reasoning over interval overlap and point containment, and order-sensitive reasoning over predecessor or successor states. Time-anchored before/after questions involve both, because they require interpreting a temporal boundary before selecting an ordered state.

\begin{figure}[t]
\centering
\includegraphics[width=0.98\columnwidth]{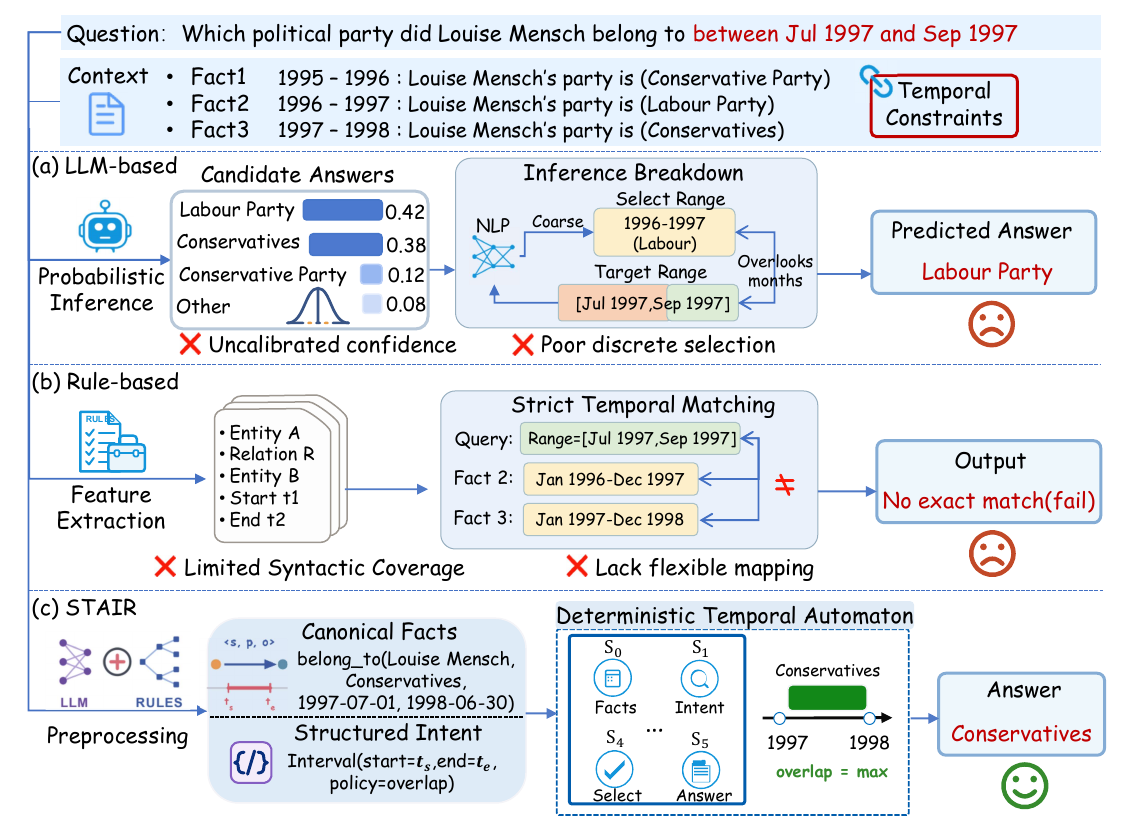}
\caption{TQA setting that motivates STAIR. Semantic parsing normalizes diverse temporal question forms, whereas deterministic execution performs temporal selection.}
\label{fig:motivation}
\end{figure}

Figure~\ref{fig:motivation} illustrates this challenge using a non-boundary-aligned interval query. An LLM can interpret the meaning of the query but may select an incorrect adjacent state through probabilistic inference. A strict rule-based matcher provides reproducible execution, yet fails when the query boundaries do not exactly align with those of the supporting fact. The underlying difficulty is thus a fundamental mismatch: flexible semantic interpretation must accommodate diverse surface forms, whereas exact temporal execution requires applying an unambiguous policy to select the correct evidence. This contrast directly motivates our approach: separating semantic normalization from deterministic temporal execution.

Recent prompting methods decompose temporal reasoning into symbolic representation, inference, verification, reflection, and answer generation. NeSTR combines symbolic temporal representations with LLM inference and feedback-based correction \citep{liang2026nestr}, while TISER constructs and revises timelines through self-reflection \citep{bazaga2025tiser}. Although these methods improve over direct prompting, the LLM still performs the decisive temporal operations and generates the final answer. Consequently, even a coherent reasoning trace may select an incorrect interval boundary, over-retrieve before/after states, or alter the answer span. Thus, we conclude the key limitation is not merely the lack of intermediate representations, but the use of free-form generation for discrete temporal decisions that admit explicit and verifiable execution.

\begin{figure}
\centering
\includegraphics[width=0.95\columnwidth]{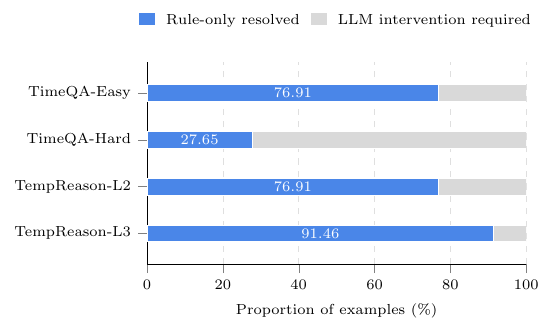}
\caption{
Routing analysis of the rule-only temporal automaton across benchmarks.
Each bar decomposes the dataset into examples deterministically resolved by rule-only automaton of STAIR and examples requiring LLM intervention.
}
\label{fig:rule_coverage}
\end{figure}

Figure~\ref{fig:rule_coverage} reveals an important asymmetry: most temporal questions already admit deterministic execution once their facts and operators are canonicalized. The rule-only path resolves 76.91\% of TimeQA-Easy, 76.91\% of TempReason-L2, and 91.46\% of TempReason-L3. These instances mainly involve explicit \texttt{from ... to ...} intervals, point-time containment, or before/after relations with entity anchors. In contrast, rule-only coverage on TimeQA-Hard is only 27.65\%, because many questions involve non-exact intervals, time-valued anchors, or formulations that do not directly expose an executable operator. For example, \texttt{between Apr 1987 and Nov 1988}, \texttt{after Jan 1996}, and \texttt{before Jan 1999} require interval-policy selection, anchor typing, or structural normalization before deterministic execution becomes applicable. These findings suggest that LLM intervention is primarily needed to interpret non-canonical temporal expressions and infer the intended temporal constraints, rather than to make every temporal decision.

Motivated by these observations, we propose \textbf{STAIR}, a Semantic-Temporal Automaton for Interpretable Reasoning. STAIR adopts a rule-first architecture separating semantic interpretation from deterministic temporal execution. It first canonicalizes temporal facts and maps recognizable questions to finite executable intents, resolved by a temporal automaton via explicit policies, finite control, and guarded transitions. If rule-based parsing fails, an answer-free semantic adapter maps the question into this intent space, while programmatic validation and constrained repair ensure executability.

Guided by dual-process reasoning \citep{kahneman2011thinking}, cognitive offloading \citep{risko2016cognitive}, and symbolic interval models \citep{allen1983maintaining}, STAIR operationalizes the principle of \emph{LLM-as-parser, Automaton-as-reasoner}. On the main reasoning path, the LLM is invoked only when difficult temporal expressions require semantic normalization, whereas the deterministic automaton performs temporal comparison, selects evidence, and extracts the answer.

Moreover, STAIR employ a procedural interpretability rather than post-hoc. For each instance resolved by the deterministic temporal automaton, the system exposes the canonical facts, normalized intent, activated temporal policy, guard outcomes, selected evidence, and answer provenance. Failures within the rule path, semantic repairs, and invocations of the final fallback are explicitly recorded. Consequently, the reasoning trace reflects the actual computation performed by the system rather than a natural-language rationale generated after prediction.

\begin{itemize}
\item We propose \textbf{STAIR}, a rule-first Semantic-Temporal Automaton for Interpretable Reasoning that instantiates the principle of \emph{LLM-as-parser, Automaton-as-reasoner} through answer-free semantic parsing and guarded deterministic temporal execution.
\item We introduce a hard-only semantic adapter that converts difficult temporal expressions, including non-exact intervals and time-anchored before/after questions, into intents that the selector can execute without allowing the LLM to choose the final answer.
\item STAIR is evaluated on TimeQA, TempReason and CronQuestions dataset.  Component-level ablations, fallback analysis, and rule-coverage diagnostics further characterize when semantic adaptation improves deterministic temporal reasoning.
\end{itemize}

\section{Related Work}

\textbf{TQA Benchmarks and Prior Systems.} Temporal reasoning has been studied through both relation-extraction resources and question-answering benchmarks. TempEval focuses on identifying temporal relations among events and time expressions \citep{verhagen2007tempeval}. TempQuestions, TimeQA, and TempReason extend this setting to question answering over time-indexed facts, evolving entities, and multi-level temporal relations \citep{jia2018tempquestions,chen2021timeqa,tan2023tempreason,tan2024complextr}. More recent benchmarks, including TRAM, TimeBench, ChronoSense, and UnSeenTimeQA, expose persistent limitations in event ordering, interval reasoning, and memorization-free temporal generalization \citep{wang2024tram,chu2024timebench,islakoglu2025chronosense,uddin2025unseentimeqa}.

Prior systems improve TQA through stronger reading-comprehension models \citep{zaheer2020bigbird,raffel2020t5,izacard2021fid}, temporal pretraining and structured knowledge representations \citep{yang2023rememo,jia2018tequila,shang2022timesensitive,mavromatis2022tempoqr}, or prompting and programmatic reasoning \citep{li2023templogic,zhu2023qaap,xiong2024tgllm,wu2024evental}. These approaches improve temporal modeling but often require supervised adaptation, task-specific graph construction, or LLM-mediated inference. STAIR instead focuses on zero-shot TQA with an explicit temporal executor.

\textbf{Inference-Time and Neuro-Symbolic Reasoning.} Inference-time reasoning methods improve LLM deliberation through intermediate rationales and sampled reasoning paths \citep{wei2022chain,wang2023selfconsistency}, search and action-based reasoning \citep{yao2023tree,yao2023react}, or feedback and additional test-time computation \citep{shinn2023reflexion,snell2025scaling,deepseekai2025r1}. In temporal QA, TISER revises constructed timelines through self-reflection, whereas NeSTR combines symbolic representations with abductive LLM reasoning. These systems organize temporal evidence and reasoning, but the LLM remains responsible for applying temporal relations and producing the final answer. Consequently, a coherent reasoning trace may still select an incorrect interval boundary or ordered state. STAIR differs by restricting the LLM to semantic normalization and executing the final temporal operation programmatically.

\textbf{Programmatic Execution and Cognitive Perspectives.} Program-aided language modeling externalizes operations that require exact and reproducible execution \citep{gao2023pal}. This division of labor is also consistent with dual-process reasoning and cognitive offloading, which motivate separating flexible interpretation from deliberate symbolic manipulation \citep{kahneman2011thinking,risko2016cognitive}. Classical interval formalisms provide a basis for precise temporal comparison \citep{allen1983maintaining}. STAIR operationalizes these perspectives through a constrained semantic interface and an explicit temporal executor.

\section{Method}
\begin{figure*}[tb]
\centering
\IfFileExists{framework.pdf}{%
    \includegraphics[width=0.98\textwidth]{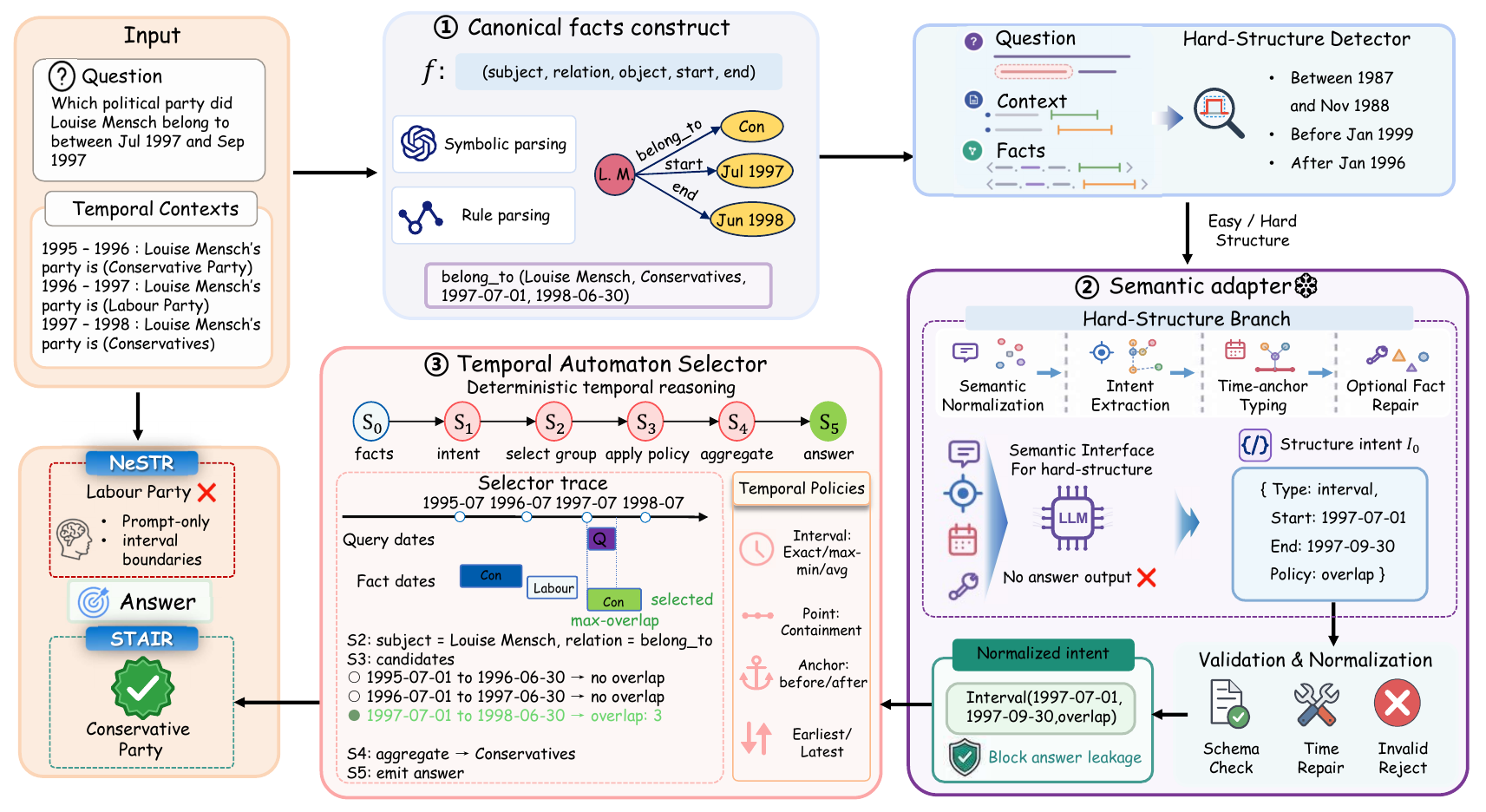}%
}{%
}
\caption{Overview of STAIR. The system first attempts rule-only execution. Difficult cases undergo semantic adaptation and validation before the Temporal Automaton Selector performs guarded temporal reasoning over canonical facts and exposes the activated policy, selected evidence, and emitted answer.}

\label{fig:framework}
\end{figure*}

Figure~\ref{fig:framework} presents an overview of STAIR. Given a question $Q$ and a temporal context $C$, the system canonicalizes the context into temporal facts, maps the question to an executable intent, and ultimately aggregates the object spans selected by the automaton to produce the final answer $A$.

The Temporal Automaton Selector (TAS) first attempts rule-only execution. If this path fails, the hard-structure detector routes supported difficult cases to an answer-free semantic adapter, whose output is validated before TAS executes again. A failed guard returns a typed reason for repair and deterministic reselection. Only when this process remains unsuccessful does STAIR invoke a \emph{4-stage LLM fallback} with separate agents for symbolic representation, temporal inference, consistency checking, and reflection/final answer generation.

\textbf{Rule-First Principle.} Rule-first execution defines STAIR's default inference regime. For directly recognizable operators such as \texttt{from t1 to t2}, \texttt{in t}, \texttt{before entity}, and \texttt{after entity}, the context is converted into canonical facts and the question is mapped to an executable intent. The automaton then applies the corresponding policy without invoking an LLM. The LLM is introduced only when the rule path fails and cannot override a successful deterministic result, making each route explicit: rule-only execution, semantic adaptation, or final fallback.

\subsection{Canonical Fact Construction}
STAIR represents each temporal fact as a canonical tuple $(s,r,o,t^s,t^e)$, where $s$, $r$, and $o$ denote the subject, relation, and object, and $t^s,t^e$ denote normalized start and end times. The system retains provenance metadata such as extraction source. Fact construction combines deterministic rule parsing with constrained LLM-based symbolic parsing. Rule parsers handle semi-structured records such as \texttt{$t_s$--$t_e$: subject's relation is object} and sentence-style descriptions such as \texttt{subject works for object from $t_s$ to $t_e$}. LLM-assisted symbolic parsing maps context statements to the same canonical schema.

The module normalizes temporal values, rejects incomplete records, and merges facts with the same normalized key while preserving their provenance metadata. These explicit representations make subsequent reasoning auditable: selected evidence can be traced to its source, and temporal policies operate over inspectable structured records rather than latent model states.

\subsection{Hard-Structure Detector and Semantic Interface}

\paragraph{Hard-Structure Detector.}
After rule-only execution fails, the Hard-structure Detector identifies cases requiring semantic normalization, including non-exact intervals (e.g., \texttt{between t1 and t2}), time-valued before/after anchors, unresolved date formats, and questions that remain unparsed despite available canonical facts. Restricting adaptation to these cases avoids unnecessary LLM calls and semantic drift.

\paragraph{Semantic Adapter.} For a difficult question $Q$, the adapter receives the question and canonical facts and produces a raw structured intent $I_0$ specifying a temporal operator, arguments, and execution policy. Supported intent types include \texttt{interval}, \texttt{point}, \texttt{before\_anchor}, \texttt{after\_anchor}, \texttt{first}, and \texttt{last}. The adapter is answer-free: it specifies execution but cannot select evidence or generate an answer.

\paragraph{Program Validation and Normalization.} We use $I_0$ for the raw adapter output and $I$ for the validated intent passed to TAS. Field aliases and date formats are canonicalized, while unsupported intents, missing temporal arguments, invalid dates, and answer-like content are rejected. Repaired facts or intents must pass the same validation procedure before deterministic reselection.

\subsection{Temporal Automaton Selector}
\paragraph{Automaton Formalization.} TAS is a guarded extended finite-state machine
\begin{equation}
\mathcal{M}=(\mathcal{S},\Xi,\delta,S_0,\mathcal{S}_{\mathrm{term}}),
\end{equation}
where $\mathcal{S}=\{S_0,\ldots,S_5,S_{\bot}\}$, $\mathcal{S}_{\mathrm{term}}=\{S_5,S_{\bot}\}$, and $\Xi$ is the configuration space. Here $S_0$ is the initial state, $S_{\bot}$ is the typed failure state, and states $S_1$ through $S_5$ record the successful completion of fact validation, intent validation, fact-group selection, policy execution, and answer aggregation, respectively. The transition function is $\delta:\mathcal{S}\times\Xi\rightarrow\mathcal{S}\times\Xi$.

A configuration is $\xi=(F,I,G,F^\star,A)$, where $F$ is the set of canonical facts, $I=(s_q,r_q,\tau,\alpha,\pi)$ is a validated intent, $s_q$ and $r_q$ are the target subject and relation, $\tau$ is the intent type, $\alpha$ is an optional entity- or time-valued anchor, $\pi$ is the execution policy, $G$ is the selected fact group, $F^\star$ is the selected evidence set, and $A$ is the answer buffer whose terminal content is returned as the final answer. For $k\in\{0,\ldots,4\}$,
\begin{equation}
\delta(S_k,\xi)=
\begin{cases}
(S_{k+1},u_k(\xi)), & g_k(\xi)=1,\\
(S_{\bot},\xi), & g_k(\xi)=0,
\end{cases}
\end{equation}
where $g_k:\Xi\rightarrow\{0,1\}$ is the guard at step $k$ and $u_k:\Xi\rightarrow\Xi$ is the corresponding deterministic update. The guards successively verify canonical facts, the validated intent, fact-group selection, policy output, and answer aggregation. A successful execution follows $S_0\!\rightarrow\!S_1\!\rightarrow\!S_2\!\rightarrow\!S_3\!\rightarrow\!S_4\!\rightarrow\!S_5$, whereas a failed guard enters $S_{\bot}$ and returns a typed failure reason to the semantic-repair controller.

The finite-state execution trace makes each prediction auditable through the evaluated guards, activated temporal policy, and selected evidence. If execution reaches $S_{\bot}$, the first failed guard identifies the stage requiring repair.

\paragraph{Fact Grouping and State Chains.} Given a set of facts $F$ and the target key $(s_q,r_q)$ from the validated intent, TAS selects
\begin{equation}
G_{s_q,r_q}=\{f_i\in F\mid s_i=s_q,\ r_i=r_q\}.
\end{equation}
Here $f_i=(s_i,r_i,o_i,t_i^s,t_i^e)$ is a canonical fact, and the selected group in the automaton configuration is $G=G_{s_q,r_q}$. Each group defines a local temporal state chain. For before/after questions, TAS follows this chain to the nearest predecessor or successor rather than returning every fact on the corresponding side of the anchor.

\paragraph{Temporal Policies.} Let $t_f^s$ and $t_f^e$ denote the normalized boundaries of a candidate fact $f$, $q=[q_s,q_e]$ an interval query with normalized boundaries $q_s$ and $q_e$, and $q_t$ a normalized point-time query. For an exact interval intent, TAS selects $F^\star=\{f\in G_{s_q,r_q}\mid t_f^s=q_s\wedge t_f^e=q_e\}$. For a non-exact interval, it computes
\begin{equation}
\omega(f,q)=\max\!\left(0,\min(t_f^e,q_e)-\max(t_f^s,q_s)\right),
\end{equation}
where $\omega(f,q)$ is the overlap length between fact $f$ and query interval $q$, and retains the facts attaining the largest positive overlap. For a point-time intent, TAS applies containment: $F^\star=\{f\in G_{s_q,r_q}\mid t_f^s\leq q_t\leq t_f^e\}$.

For an entity-anchored intent, TAS locates the anchor fact $f_a$, whose normalized boundaries are $t_a^s$ and $t_a^e$, and selects the nearest temporal predecessor or successor:
\begin{equation}
F^\star=
\begin{cases}
\displaystyle \arg\max_{f\in G_{s_q,r_q}:\,t_f^e\leq t_a^s} t_f^e, & \text{before},\\[3pt]
\displaystyle \arg\min_{f\in G_{s_q,r_q}:\,t_f^s\geq t_a^e} t_f^s, & \text{after}.
\end{cases}
\end{equation}
For time-valued anchors, the same predecessor or successor search is applied directly to the normalized anchor time. For \texttt{first} and \texttt{last} intents, TAS selects the earliest and latest facts in the corresponding temporal state chain, respectively. If a policy yields no admissible evidence or multiple distinct answers, the guard at $S_3$ routes the instance to $S_{\bot}$ with a typed failure reason.

\subsection{Direct Answer Emission and Fallback}
When TAS reaches $S_5$, STAIR copies the answer directly from the selected object spans, preventing unsupported entities and surface-form changes. If a guard fails, the returned failure type is used to repair the canonical facts or intent without generating an answer. Every repaired output must be revalidated before TAS is executed again.

STAIR invokes the 4-stage LLM fallback only when semantic repair and deterministic reselection both fail. Unlike the answer-free repair branch, this final path may generate an answer through its reflection and final-answer stage, preserving coverage for cases outside the current policy inventory.

\begin{table*}[!tb]
\centering
\small % 在 10pt 主字体下，\small 对应 9pt 
\setlength{\tabcolsep}{5.85pt}
\begin{tabular}{>{\centering\arraybackslash}m{0.9cm}|l|lcccccccccc}
\toprule

\multicolumn{2}{c}{\multirow{2}{*}{Model}}
& \multirow{2}{*}{Strategy}
& \multicolumn{2}{c}{TimeQA-Easy}
& \multicolumn{2}{c}{TimeQA-Hard}
& \multicolumn{2}{c}{TempReason-L2}
& \multicolumn{2}{c}{TempReason-L3}
& \multicolumn{2}{c}{Avg} \\
\cmidrule(lr){4-5}
\cmidrule(lr){6-7}
\cmidrule(lr){8-9}
\cmidrule(lr){10-11}
\cmidrule(lr){12-13}
\multicolumn{2}{c}{} 
& & EM & F1 & EM & F1 & EM & F1 & EM & F1 & EM & F1 \\
\hline

\multirow[c]{9}{*}{\shortstack{Open\\LLMs}}
& \multirow{3}{*}{Qwen2.5-7B}
& \rule{0pt}{2.6ex}TISER & 86.80 & 92.60 & 64.30 & 71.50 & 61.10 & 69.80 & 72.60 & 77.60 & 71.20 & 77.90 \\ % 【修复2】加入顶端隐形支架
& & NeSTR & 85.10 & 90.20 & 64.80 & 71.20 & 61.50 & 68.60 & 73.10 & 76.70 & 71.10 & 76.70 \\
& & \textbf{STAIR}
& 93.27 & 94.25
& 76.94 & 80.91
& 87.61 & 87.72
& 94.90 & 94.77
& 88.18 & 89.41 \\[0.8ex] % 【修复3】强制本行底部增加 0.8ex 间距
\cline{2-13}

& \multirow{3}{*}{Qwen3-8B}
& \rule{0pt}{2.6ex}TISER & 88.80 & 93.40 & 77.10 & 82.50 & 73.70 & 78.40 & 84.30 & 87.50 & 80.90 & 85.40 \\
& & NeSTR & 89.50 & 94.20 & 77.70 & 83.40 & 79.20 & 83.50 & 84.90 & 87.20 & 82.80 & 87.10 \\
& & \textbf{STAIR}
& 94.79 & 95.52
& \textbf{84.32} & 86.37
& 88.75 & \textbf{91.31}
& \textbf{96.18} & 95.52
& 91.01 & 92.18 \\[0.8ex]
\cline{2-13}

& \multirow{3}{*}{Qwen3-14B}
& \rule{0pt}{2.6ex}TISER & 90.00 & 94.30 & 82.10 & 87.20 & 75.50 & 80.60 & 81.60 & 85.20 & 82.30 & 86.80 \\
& & NeSTR & 91.10 & 94.50 & 82.20 & \textbf{87.30} & 79.50 & 84.60 & 85.10 & 88.90 & 84.50 & 88.80 \\
& & \textbf{STAIR}
& 95.01 & 96.07
& 84.18 & 86.59
& 86.69 & 90.87
& 95.96 & 95.44
& 90.46 & 92.25 \\[0.8ex]
\hline

\multirow[c]{3}{*}{\shortstack{Closed\\LLMs}}
& \multirow{3}{*}{GPT-4o-mini}
& \rule{0pt}{2.6ex}TISER & 86.70 & 91.90 & 74.30 & 79.90 & 77.70 & 84.10 & 82.30 & 87.10 & 80.20 & 85.80 \\
& & NeSTR & 93.70 & 96.40 & 81.70 & 85.90 & 80.80 & 86.40 & 84.60 & 90.00 & 85.20 & 89.70 \\
& & \textbf{STAIR}
& \textbf{96.57} & \textbf{97.02}
& 83.97 & 86.24
& \textbf{89.13} & 91.10
& 96.16 & \textbf{95.54}
& \textbf{91.46} & \textbf{92.48} \\[0.8ex]

\hline
\end{tabular}

\caption{Exact Match (EM) and token-level F1 results on four temporal reasoning benchmarks under matched model settings.All four benchmarks are evaluated on their complete test sets. NeSTR and TISER values are taken from prior work, while STAIR values are averaged over three independent runs. Boldface indicates the best result for each model and metric.}
\label{tab:main}
\end{table*}

\section{Experiments}

\subsection{Experimental Setup}
\paragraph{Datasets.}
We evaluate STAIR on the complete test sets of TimeQA-Easy and TimeQA-Hard from TimeQA, as well as TempReason-L2 and TempReason-L3 from TempReason. We additionally evaluate cross-source transfer on the operator-supported subset of the official CronQuestions test split \citep{saxena2021cronquestions}, converted to the same question--context format.
\paragraph{Baselines and Models.}
The primary comparison is with NeSTR under matched model and benchmark settings. Table~\ref{tab:main} reports the published NeSTR scores; we separately reproduced its original structured prompt and obtained closely aligned results, using this reproduction only as a consistency check. TISER is included as an external matched-model reference using values reported by its authors. We evaluate Qwen2.5-7B, Qwen3-8B, Qwen3-14B, and GPT-4o-mini.

\paragraph{Implementation Details.}
All STAIR runs use the same input contexts, data splits, answer normalization procedure, and evaluation metrics as the NeSTR comparison. For the 4-stage LLM fallback, STAIR adapts the original NeSTR stage descriptions into four agent-specific prompts, invoked only when both semantic repair and deterministic reselection fail. This decomposition is applied only to STAIR; NeSTR remains unmodified. Generation uses temperature $0.1$ and a maximum output length of 1024 tokens. STAIR is run independently three times, and we report mean performance in the main paper. Standard deviations are provided in the Appendix. No manual filtering is applied to the four main benchmarks.

The experiments are conducted in a PyTorch 2.11.0 with CUDA 13.0 environment running on Ubuntu 24.04, with an NVIDIA GeForce RTX 4090 GPU used for acceleration.
\paragraph{Evaluation Metrics.}
We report Exact Match (EM) and token-level F1 following the standard TQA protocol. EM requires the normalized prediction to match the reference exactly, while F1 assigns partial credit through token overlap, distinguishing exact entity or event selection from partially correct spans.

\subsection{Main Results}
Table~\ref{tab:main} demonstrates that STAIR achieves the highest average Exact Match (EM) and F1 scores across all evaluated models. Relative to NeSTR, STAIR increases the average F1 score by margins ranging from 2.78 points (with GPT-4o-mini) to 12.71 points (with Qwen2.5-7B), accompanied by EM gains between 5.96 and 17.08 points. Across 16 distinct model-dataset configurations, STAIR improves EM in all cases and F1 in 15. Furthermore, the average F1 score of STAIR varies by only 3.07 points across different models, compared with a variance of 13 points for NeSTR. This significant reduction in variance demonstrates a decreased sensitivity to the underlying capacity of the language model.

The most substantial improvements occur on TempReason-L2 and TempReason-L3, achieving average F1 enhancements of 9.48 and 9.62 points over NeSTR, respectively. In contrast, the corresponding gains are 1.89 points on TimeQA-Easy and 3.08 points on TimeQA-Hard. This pattern aligns with the underlying temporal operations. TempReason-L2 and TempReason-L3 primarily contain point-time containment and entity-anchored before/after queries, for which boundary- and order-sensitive decisions map directly to deterministic policies. TimeQA-Hard instead concentrates non-exact intervals and time-anchored before/after queries, which require semantic normalization before the same policies can be executed. Notably, EM improvements consistently surpass F1 gains, indicating that STAIR enhances exact evidence selection rather than merely inflating partial lexical overlap.

While performance remains consistent overall, the most notable variations emerge within TimeQA-Hard. For instance, when utilizing the Qwen3-14B model on this subset, STAIR increases the EM score from 82.20 to 84.18 but simultaneously experiences a marginal F1 decrease from 87.30 to 86.59. This localized exception highlights a fundamental and inherent trade-off: imposing deterministic exact selection can occasionally penalize granular token-level overlap on complex temporal queries where standard baseline models might generate partially correct, yet overly verbose responses.

\subsection{Component-level Ablation}
\begin{table}
\centering
\small
\setlength{\tabcolsep}{2.0pt}
\begin{tabular}{lccccc}
\toprule
Variant & EM & F1  & Adapter & Fallback (\%) \\
\midrule
STAIR-Core & 68.06 & 72.62  & -- & 72.35 \\
+ hard detector & 68.29 & 72.75  & -- & 72.35 \\
+ semantic adapter & 74.95 & 80.91  & 35.35 & 37.00 \\
+ validation/repair & 75.11 & 81.12  & 35.48 & 36.87 \\
STAIR complete & \textbf{77.16} & \textbf{81.20}  & 43.73 & 20.50 \\
w/o max-overlap & 74.01 & 80.26  & 43.53 & 20.89 \\
w/o time-anchor typing & 76.28 & 80.19  & 36.39 & 28.01 \\
w/o 4-stage lLM fallback & 70.01 & 72.47 & 43.47 & -- \\
\bottomrule
\end{tabular}
\caption{Interface ablation on TimeQA-Hard with Qwen2.5-7B. Fallback denotes the final 4-stage LLM fallback rate.}
\label{tab:interface_ablation}
\end{table}

\paragraph{Independent-Run Variation.}
The Qwen2.5-7B STAIR entries in Tables \ref{tab:main} and \ref{tab:interface_ablation} are evaluated under the identical running environment. The minor differences in EM and F1 arise from sampling at temperature 0.1, not from a change in configuration. We use TimeQA-Hard for the main ablation study because its non-exact intervals, time-valued anchors, and low rule-only coverage make it the most diagnostic setting for evaluating STAIR.

Table~\ref{tab:interface_ablation} deconstructs the architecture of STAIR on the TimeQA-Hard dataset to evaluate the individual contributions of the core modules. Note that executor-side ablations are deferred to the Appendix. 

The result shows that the hard-structure detector alone provides little benefit because detection does not resolve the underlying interface failure. The main improvement comes from the semantic adapter, which converts previously unsupported question forms into executable intents, increasing F1 from 72.75 to 80.91 while reducing the final fallback rate from 72.35\% to 37\%. Validation and repair yield only a modest additional accuracy gain, but they improve reliability by preventing malformed intents from entering the automaton.

The complete system achieves the best EM and F1 together with the lowest fallback rate. The policy ablations reveal two distinct effects: removing max-overlap primarily degrades evidence selection quality, whereas removing time-anchor typing reduces the proportion of questions that can be executed deterministically. Eliminating the 4-stage LLM fallback causes a substantial performance drop, confirming its role as a coverage mechanism for cases outside the current policy inventory. Overall, the results support the intended division of labor: the LLM maps difficult language into constrained structures, while the automaton performs the final temporal decision through explicit policies.

\subsection{Cross-source Transfer Assessment}
Table~\ref{tab:cron} evaluates transfer on the successfully converted, operator-supported subset of CronQuestions. From the official test split of 30,000 instances, we exclude 14,308 questions with unsupported task or answer types and 2,596 type-supported questions whose annotations lack the \texttt{head} field required by the converter to retrieve a subject--relation temporal timeline. This yields 13,096 instances covering \texttt{simple\_entity}, entity-answer \texttt{first\_last}, and entity-answer \texttt{before\_after}. NeSTR and STAIR use exactly the same converted questions, answers, and temporal contexts. 

\begin{table}
\centering
\small
\begin{tabular}{lccc}
\toprule
Method & Samples & EM & F1 \\
\midrule
NeSTR & 13,096 & 72.00 & 81.03 \\
STAIR & 13,096 & 88.42 & 87.63 \\
Improvement & -- & +16.42 & +6.60 \\
\bottomrule
\end{tabular}
\caption{Cross-source transfer results on the operator-supported CronQuestions subset. All model configurations are evaluated over three independent runs.}
\label{tab:cron}
\end{table}

On this shared subset, STAIR outperforms NeSTR by 16.42 EM points and 6.60 F1 points. The result demonstrates transfer of the current canonical representation and temporal policy inventory across data sources. It should be read as evidence for the converted, operator-supported subset rather than as a claim about unseen temporal operators or the complete CronQuestions test split.

\subsection{Efficiency and Fallback Analysis}
As shown in Table~\ref{tab:efficiency}, The efficiency pattern is structure dependent. On TimeQA-Easy, TempReason-L2, and TempReason-L3, STAIR reduces calls, token usage, and wall-clock time; the largest reduction appears on TempReason-L3, where average calls decrease from 1 to 0.21 and time from 3.90s to 0.31s. On TimeQA-Hard, non-exact intervals and time-anchored before/after questions require additional adaptation and repair, increasing calls from 1 to 2.12 and wall-clock time from 3.13s to 3.61s. This result reflects an accuracy-efficiency trade-off on structurally difficult inputs: STAIR spends additional computation to convert complex language into validated intents while preserving deterministic final selection. 

\begin{table}
\centering
\small
\setlength{\tabcolsep}{1.85pt}
\begin{tabular}{llccccc}
\toprule
Dataset & System & Calls & Input tok. & Output tok. & Time \\
\midrule
\multirow{2}{*}{TimeQA-Easy}
 & NeSTR & 1.00 & 548.0 & 288.1 & 2.66s \\
 & STAIR & 0.64 & 256.0 & 52.9  & 0.82s \\
\midrule
\multirow{2}{*}{TimeQA-Hard}
 & NeSTR & 1.00 & 543.8 & 351.5 & 3.13s \\
 & STAIR & 2.12 & 929.3 & 282.0 & 3.61s \\
\midrule
\multirow{2}{*}{TempReason-L2}
 & NeSTR & 1.00 & 689.6 & 462.2 & 3.98s \\
 & STAIR & 0.89 & 545.8 & 100.4 & 1.44s \\
\midrule
\multirow{2}{*}{TempReason-L3}
 & NeSTR & 1.00 & 707.3 & 450.3 & 3.90s \\
 & STAIR & 0.21 & 81.9  & 19.8  & 0.31s \\
\midrule
\multicolumn{2}{l}{Macro summary}
 & $-3.65\%$ & \multicolumn{2}{c}{tokens: $-43.87\%$}
 & $2.21\times$ \\
\bottomrule
\end{tabular}
\caption{Efficiency comparison between the NeSTR baseline and STAIR.  Calls and token counts are reported per instance. Results from a single run of Qwen2.5-7B-Instruct.}
\label{tab:efficiency}
\end{table}

\section{Diagnostic Analysis}
\paragraph{Category-level Question Structure.}
Table~\ref{tab:category_diagnostic} operationalizes the two high-level temporal challenges into four executable query categories. Non-exact interval and point-time questions instantiate boundary-sensitive reasoning through interval overlap and point containment, respectively. Time-anchored before/after questions combine boundary-sensitive anchor interpretation with order-sensitive predecessor or successor selection.

STAIR achieves F1 gains of 8.19 points on non-exact intervals and 9.28 points on point-time queries. The former relies strongly on semantic adaptation, whereas the latter is handled without adapter intervention, showing that deterministic execution is beneficial both after semantic normalization and when the temporal operator is explicit. For time-anchored queries, STAIR improves F1 by 6.92 points on before questions and 4.56 points on after questions. The substantially higher fallback rate for time-anchor after indicates that this category more frequently exceeds the coverage of the current semantic interface and policy inventory.

\begin{table}[t]
\centering
\scriptsize
\setlength{\tabcolsep}{2.0pt}
\renewcommand{\arraystretch}{1.08}

% Panel (a): performance and improvements
\begin{tabular*}{\columnwidth}{
@{\extracolsep{\fill}}
>{\raggedright\arraybackslash}p{1.85cm}
c
r
r r
r r
r r
@{}
}
\toprule
\multicolumn{9}{@{}l}{\textit{(a) Performance by temporal category}} \\[-1pt]

\multirow{2}{*}{Category}
& \multirow{2}{*}{Property}
& \multirow{2}{*}{$n$}
& \multicolumn{2}{c}{NeSTR}
& \multicolumn{2}{c}{STAIR}
& \multicolumn{2}{c}{$\Delta$} \\
\cmidrule(lr){4-5}
\cmidrule(lr){6-7}
\cmidrule(l){8-9}

& & & EM & F1 & EM & F1 & EM & F1 \\
\midrule

Non-exact interval
& B
& 1285
& 68.53 & 75.02
& 78.60 & 83.21
& +10.06 & +8.20 \\

Point-time
& B
& 1411
& 62.37 & 69.26
& 74.34 & 78.54
& +11.98 & +9.28 \\

Time-anchor before
& B+O
& 248
& 76.88 & 80.57
& 87.50 & 87.49
& +10.62 & +6.93 \\

Time-anchor after
& B+O
& 134
& 65.17 & 73.61
& 73.88 & 78.17
& +8.71 & +4.56 \\

\end{tabular*}

% Panel (b): execution diagnostics
\begin{tabular*}{\columnwidth}{
@{\extracolsep{\fill}}
>{\raggedright\arraybackslash}p{2.65cm}
r
r
@{}
}
\toprule
\multicolumn{3}{@{}l}{\textit{(b) STAIR execution diagnostics}} \\[-1pt]

Category
& Adapter (\%)
& Fallback (\%) \\
\midrule

Non-exact interval
& 86.15
& 13.62 \\

Point-time
& 0.00
& 22.18 \\

Time-anchor before
& 83.47
& 16.53 \\

Time-anchor after
& 23.88
& 76.12 \\

\bottomrule
\end{tabular*}

\begingroup
\raggedright
\scriptsize
\par
\endgroup

\caption{
Single-run category-level diagnostics on TimeQA-Hard using
Qwen2.5-7B-Instruct. B denotes boundary-sensitive, whereas B+O denotes both boundary-sensitive and order-sensitive. $\Delta$ is calculated as STAIR minus NeSTR.
}
\label{tab:category_diagnostic}
\end{table}

\paragraph{Error Analysis.}
The majority of residual errors arise when contexts resist conversion into canonical facts, query boundaries intersect multiple plausible intervals, adapter outputs fail validation, or questions require operators outside the predefined policy inventory. Currently unsupported phenomena include highly implicit event ordering, duration comparisons, negation, nested constraints, and multi-hop temporal composition. These limitations reflect the deterministic design scope of STAIR, which requires canonical facts, a finite temporal intent, and directly extractable answers. Instances violating these assumptions trigger rejection; if repair processes fail, they are routed to the LLM fallback mechanism.

Consequently, while this fallback expands the range of addressable queries, these predictions inherently lack the guard-level execution traces provided by the deterministic TAS module, causing a loss of procedural interpretability.

\section{Conclusion}
We presented STAIR, a rule-first framework separating semantic interpretation from precise temporal execution in zero-shot TQA. Regular questions are resolved directly by a deterministic temporal automaton, while an answer-free semantic adapter maps difficult formulations into validated intents executable by the same automaton over canonicalized evidence. Experiments on four TQA benchmarks and the operator-supported subset of CronQuestions demonstrate consistent improvements over the NeSTR baseline across open-source and proprietary models. Category-level diagnostics show gains on non-exact intervals, point-time containment, and time-anchored before/after questions, covering both boundary- and order-sensitive reasoning. Ablations attribute these improvements to semantic adaptation, time-anchor typing, guarded temporal policies, and deterministic answer emission. Efficiency results show that rule-first execution substantially reduces model calls on structurally regular datasets, although difficult constraints incur additional adaptation costs. STAIR demonstrates that restricting LLMs to semantic interfaces while delegating discrete temporal decisions to interpretable executors provides an effective and reliable approach to temporal question answering.

\bibliography{references}
\clearpage
\includepdf[pages=-,pagecommand={\thispagestyle{empty}}]{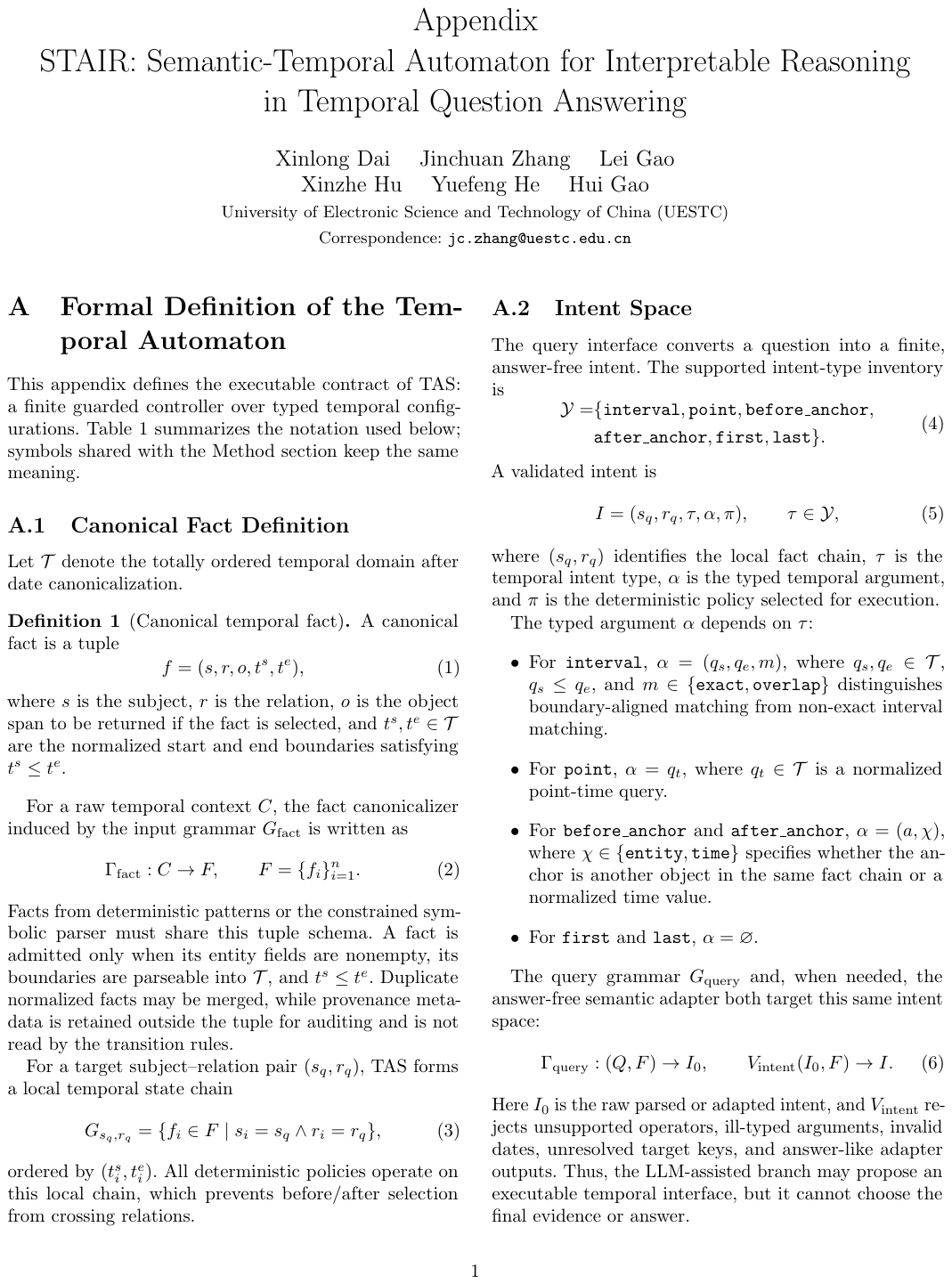}
\end{document}